\documentclass[runningheads]{llncs}
\usepackage[T1]{fontenc}
\usepackage{graphicx,verbatim}
\usepackage{multirow}
\usepackage[table,xcdraw]{xcolor}
\usepackage{longtable}
\newlength\savewidth\newcommand\shline{\noalign{\global\savewidth\arrayrulewidth
  \global\arrayrulewidth 1.25pt}\hline\noalign{\global\arrayrulewidth\savewidth}}
\usepackage{booktabs}
\usepackage{nicematrix} 
\newcommand{\eg}{\textit{e}.\textit{g}.}
\newcommand{\chwang}[1]{#1}
\usepackage{xspace}
\usepackage[colorlinks=true,citecolor=blue,linkcolor=blue,urlcolor=blue]{hyperref}

\newcommand{\methodname}{{UniField}\xspace}
\begin{document}
\title{UniField: A Unified Field-Aware MRI Enhancement Framework}

\author{
    Anonymized Authors\footnotemark[1]\footnotemark[2] \\
    Anonymized Affiliations \\
    email@anonymized.com
}

\renewcommand{\thefootnote}{\fnsymbol{footnote}}
\footnotetext[1]{Yiyang Lin and Chenhui Wang contribute equally to this work.}
\footnotetext[2]{Corresponding author: Yixuan Yuan.}
\author{Yiyang Lin\inst{1} \and
Chenhui Wang\inst{2} \and
Zhihao Peng\inst{1} \and
Yixuan Yuan\inst{1}}
\authorrunning{F. Author et al.}
\institute{Department of Electronic Engineering, the Chinese University of Hong Kong \and
The Institute of Science and Technology for Brain-inspired Intelligence, Fudan University\\
\email{yylin247@163.com, chenhuiwang21@m.fudan.edu.cn, yxyuan@ee.cuhk.edu.hk}}

\maketitle              % typeset the header of the contribution
\begin{sloppypar}
\begin{abstract}

Magnetic Resonance Imaging (MRI) field-strength enhancement holds immense value for both clinical diagnostics and advanced research. However, existing methods typically focus on isolated enhancement tasks, such as specific 64mT-to-3T or 3T-to-7T transitions using limited subject cohorts, thereby failing to exploit the shared degradation patterns inherent across different field strengths and severely restricting model generalization. To address this challenge, we propose \methodname, a unified framework integrating multiple modalities and enhancement tasks to mutually promote representation learning by exploiting these shared degradation characteristics. Specifically, our main contributions are threefold. Firstly, to overcome MRI data scarcity and capture continuous anatomical structures, \methodname departs from conventional methods that treat 3D MRI volumes as independent 2D slices. Instead, we directly exploit comprehensive 3D volumetric information by leveraging pre-trained 3D foundation models, thereby embedding generalized and robust structural representations to significantly boost enhancement performance. In addition, to mitigate the spectral bias of mainstream flow-matching models that often over-smooth high-frequency details, we explicitly incorporate the physical mechanisms of magnetic fields to introduce a Field-Aware Spectral Rectification Mechanism (FASRM), tailoring customized spectral corrections to distinct field strengths. Finally, to resolve the fundamental data bottleneck, we organize and publicly release a comprehensive paired multi-field MRI dataset, which is an order of magnitude larger than existing datasets. Extensive experiments demonstrate our method's superiority over state-of-the-art approaches, achieving an average improvement of approximately 1.81 dB in PSNR and 9.47\% in SSIM. Codes and datasets are available at: https://github.com/linyiyang98/UniField.

% Authors must provide keywords and are not allowed to remove this Keyword section.

\end{abstract}
\section{Introduction}
Magnetic Resonance Imaging (MRI) \cite{katti2011magnetic,peng2025omnibrainbench,wang2026brain} is crucial for medical diagnosis, particularly in brain imaging.
% Renowned for its exceptional soft-tissue contrast and non-invasive nature, MRI serves as the gold standard for detecting complex neurological conditions such as tumors, acute strokes, and neurodegenerative diseases. 
In clinical practice, MRI utilizes various field strengths\chwang{: ultra-low-field (64mT) scanners offer portability and accessibility, standard clinical-field (3T) provides high-resolution structural mapping, and ultra-high-field (7T) reveals unprecedented microscopic details.}
% , for instance, ultra-low field (64mT) offers portability, high field (3T) provides standard high-resolution, and ultra-high field (7T) reveals microscopic details. 
% To combine the portability and low cost of low-field systems with the clarity of high-field ones, computational enhancement techniques can convert low-field MRI into high-field quality. For instance, converting 64mT images to 3T quality enables portable devices to achieve 3T-level accuracy, allowing bedridden patients unable to visit hospitals to receive high-quality diagnostics directly at their bedside.
\chwang{Computational field-strength enhancement techniques offer a transformative solution to bridge these hardware gaps. Upgrading low-cost 64mT scans to 3T quality democratizes high-fidelity diagnostics at the patient's bedside, while enhancing clinical 3T images to 7T quality unlocks advanced neuroimaging capabilities without the exorbitant costs of ultra-high-field systems \cite{he2024f2tnet,peng2024gbt}.}

\begin{figure}[t]
    \centering
    % 如果图片太大或太小，可以调整 width 的值，例如 width=0.8\linewidth 或 width=8cm
    \includegraphics[width=\linewidth]{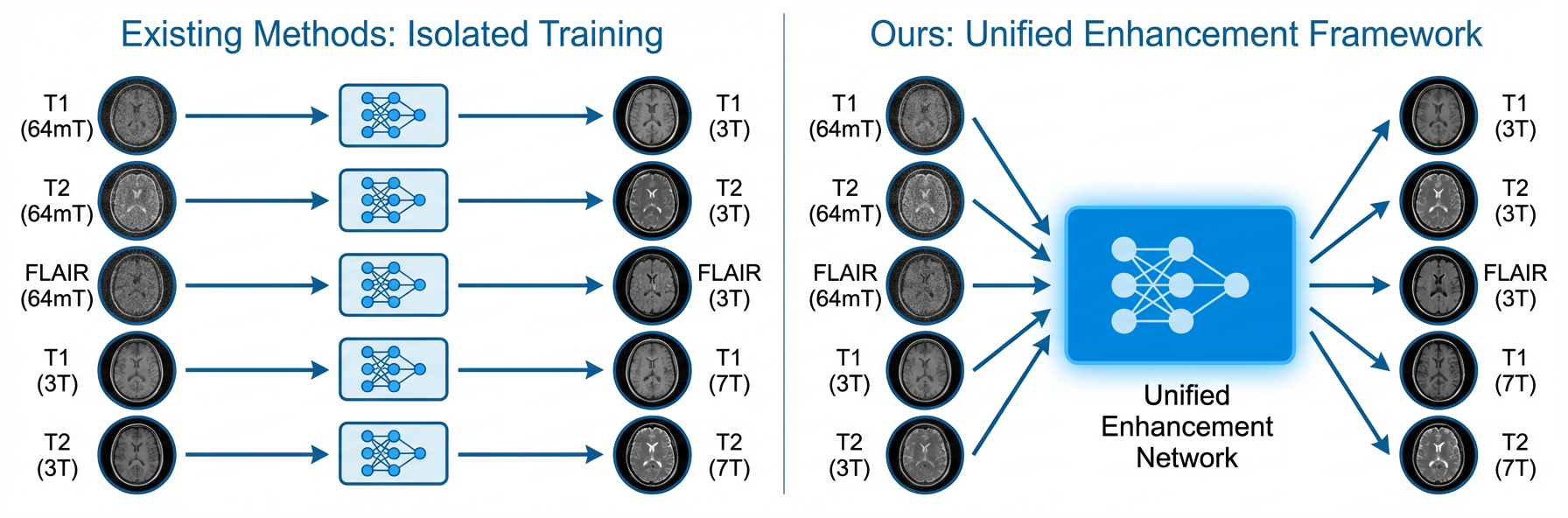}
    \caption{\textbf{Comparison of existing MRI enhancing method and our framework.} (Left) Existing methods train isolated networks for specific modalities and tasks. (Right) Our method unifies all modalities and tasks within a single framework.}
    \label{fig:intro}
\end{figure}

However, developing robust computational MRI enhancement models is severely bottlenecked by three main challenges. First, the fundamental data scarcity and isolated training paradigms remain major hurdles. Existing studies typically rely on merely a few dozen strictly paired cases \cite{chen2023paired,chu2025paired,vasa2025ultra}, and predominantly train isolated models for specific modalities (\eg~T1, T2) or localized tasks (\eg~64mT-to-3T). This fragmented approach inevitably causes overfitting (Fig. \ref{fig:intro}, left) and fails to exploit the shared degradation patterns inherent across different field strengths \cite{fei2023generative,yang2024crnet}. Second, conventional methods often treat 3D MRI volumes as independent 2D slices. This slice-by-slice processing discards crucial inter-slice continuous anatomical structures, severely limiting structural fidelity. Third, mainstream MRI enhancement methods, such as flow-matching based models, suffer from severe spectral bias. By overlooking the distinct physical mechanisms of magnetic fields at varying strengths, they tend to over-smooth high-frequency details that are vital for clinical diagnosis.

To address these intertwined challenges, we propose \methodname, a unified framework that integrates multiple modalities and enhancement tasks to mutually promote representation learning. Specifically, our solutions are threefold, directly targeting the aforementioned bottlenecks. Firstly, to resolve the fundamental data scarcity and isolated training limitations, we organize and publicly release a comprehensive paired multi-field MRI dataset \cite{wang2024joint,wang2024hope}, which is an order of magnitude larger than existing benchmarks, enabling our unified model to fully exploit shared degradation characteristics. In addition, to overcome the limitations of 2D processing, \methodname departs from conventional slice-by-slice methods. Instead, we exploit comprehensive 3D volumetric information by leveraging pre-trained 3D foundation models, thereby embedding generalized and robust structural representations. Finally, to mitigate the spectral bias and preserve fine anatomical details, we explicitly incorporate the physical mechanisms of magnetic fields to introduce a Field-Aware Spectral Rectification Mechanism (FASRM), tailoring customized spectral corrections to distinct field strengths.

\chwang{
Our main contributions can be summarized as follows:
(1) We propose UniField to unify diverse MRI modalities and enhancement tasks, leveraging shared intrinsic degradation patterns for mutual performance promotion.
(2) To mitigate the spectral bias and over-smoothing inherent in traditional flow-matching, we introduce a novel dual-domain paradigm. It leverages the decoupled frequency domain to tailor customized spectral optimization schemes based on the distinct physical characteristics of varying field strengths.
(3) We curate and release the largest paired multi-field MRI enhancement dataset to date. By precisely registering previously unaligned data, we increase benchmark volumes by an order of magnitude, establishing a robust foundation for future research.
}
% In summary, we propose the first unified MRI field-strength enhancement model. Our main contributions are threefold: (1) We unify diverse MRI modalities and enhancement tasks within a single framework, leveraging shared intrinsic degradation patterns to enable mutual performance promotion and achieve superior results for each specific task; (2) We propose a Physics-Guided Dual-Domain Flow-Matching framework that tailors customized spectral optimization schemes for each task in the frequency domain, effectively resolving the inherent spectral bias and over-smoothing issues of traditional flow-matching; and (3) We curate a large-scale MRI enhancement dataset by precisely registering all previously unaligned data, increasing the volume of existing benchmarks by an order of magnitude, which will be made publicly available to establish a robust foundation for future research.

\section{Method}
\subsection{Overall Architecture of UniField}

\begin{figure}[t]
    \centering
    % 如果图片太大或太小，可以调整 width 的值，例如 width=0.8\linewidth 或 width=8cm
    \includegraphics[width=\linewidth]{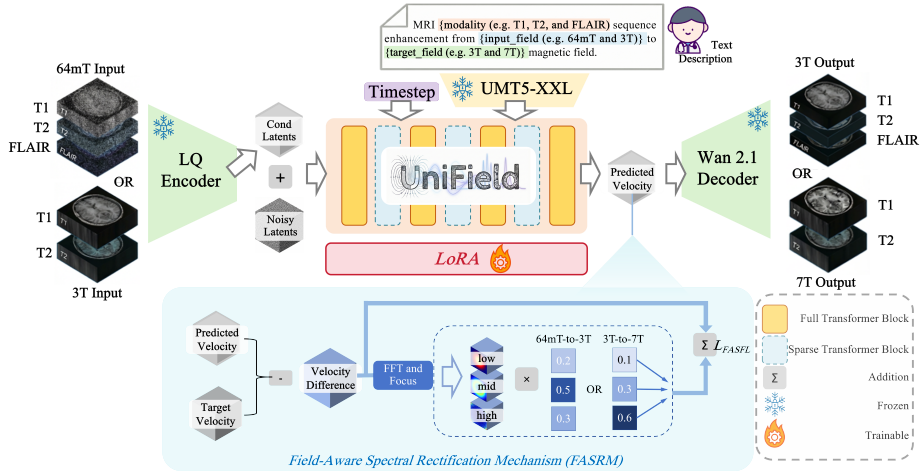}
    \caption{\textbf{Overview of our proposed UniField.} Within the framework, the encoder and decoder utilize frozen pre-trained FlashVSR weights, while the central unified network is fine-tuned using LoRA.}
    \label{fig:method}
\end{figure}

\chwang{To overcome the data scarcity and poor generalization of isolated models, we propose UniField, a unified MRI field-strength enhancement framework (Fig. \ref{fig:method}). This unified framework integrates diverse modalities (\eg~T1, T2, FLAIR) and cross-field enhancements---such as low-field (LF) 64\text{mT} to high-field (HF) 3\text{T}, and 3\text{T} to 7\text{T}---to mutually promote performance via shared degradation patterns. Furthermore, to embed robust structural priors, UniField operates within the latent space of FlashVSR \cite{flashvsr}, a pre-trained video super-resolution model.}

\chwang{During the forward process, flow-matching constructs a time-dependent latent state $\mathbf{z}_t$ at $t \in [0, 1]$ by interpolating between the clean HF target latent $\mathbf{z}_0 = \mathcal{E}(\mathbf{x}_{\text{HF}})$ and standard Gaussian noise $\mathbf{z}_1 \sim \mathcal{N}(\mathbf{0}, \mathbf{I})$. The core network $\mathcal{F}_{\theta}$ is trained to predict the velocity field driving this transition, explicitly conditioned on the current timestep $t$.}

\chwang{During inference, given an LF input $\mathbf{x}_{\text{LF}}$, initial noise $\mathbf{z}_1$, and a frozen UMT5-XXL \cite{pirhadi2025cvt5}-encoded text condition $\mathbf{c}$ (formulated as ``MRI \{modality (e.g., T1, T2, and FLAIR)\} sequence enhancement from \{input\_field (e.g., 64mT and 3T)\} to \{target\_field (e.g., 3T and 7T)\} magnetic field''), we solve the empirical ODE:}

\begin{equation}
    \hat{\mathbf{x}}_{\text{HF}} = \mathcal{D}\left( \text{ODE}(\mathcal{F}_{\theta}(\mathbf{z}_1, \mathcal{E}(\mathbf{x}_{\text{LF}}), \mathbf{c})) \right),
\end{equation}
\chwang{where $\mathcal{E}(\cdot)$ and $\mathcal{D}(\cdot)$ are the frozen FlashVSR encoder and decoder. Specifically, the ODE solver $\text{ODE}(\cdot)$ numerically evaluates the time-dependent predicted velocity field $\mathcal{F}_{\theta}$ across timesteps from $t=1$ to $t=0$, yielding a clean latent representation $\mathbf{z}_0$ that $\mathcal{D}(\cdot)$ reconstructs into the target HF volume $\hat{\mathbf{x}}_{\text{HF}}$. To efficiently adapt $\mathcal{F}_{\theta}$ for 3D MRI, we employ Low-Rank Adaptation (LoRA) \cite{sun2022recent} alongside sparse attention \cite{sun2022recent}, preventing long-range spatial dependencies from blurring critical local anatomy.}

 \chwang{However, while this latent flow-matching process provides a powerful generative backbone, it inherently suffers from spectral bias, often resulting in over-smoothed structures. To mitigate this, we introduce a Field-Aware Spectral Rectification Mechanism (FASRM) (Fig. \ref{fig:method}, blue box), which explicitly optimizes high-frequency details based on field-specific physical characteristics.
 }
% where $\mathcal{E}$ and $\mathcal{D}$ denote the frozen video-prior FlashVSR encoder and decoder , respectively, and $\mathcal{T}$ represents the frozen UMT5-XXL text encoder used to extract task-specific conditions. 
% The central unified network, $\mathcal{F}_{\theta}$, predicts the velocity field from sampled Gaussian noise $z_t$ to construct a deterministic generation trajectory. To adapt this video-based architecture to 3D MRI efficiently, $\mathcal{F}_{\theta}$ is fine-tuned using Low-Rank Adaptation (LoRA). Additionally, it incorporates sparse attention mechanisms to prevent irrelevant long-range spatial interactions from overwhelming critical local anatomical details. 
% Finally, while latent flow-matching provides a powerful generative backbone, it inherently suffers from spectral bias. To mitigate the resulting over-smoothed structures, we introduce a Physics-Guided Dual-Domain Flow-Matching module (Fig. \ref{fig:method}, blue box). }
% where \(\mathcal{E}\) and \(\mathcal{D}\) denote the frozen pre-trained encoder and decoder respectively, and \(\mathcal{T}\) is the frozen UMT5-XXL text encoder. \(\mathcal{F}_{\theta}\) is the central unified network (fine-tuned via LoRA) that predicts the velocity field to construct a deterministic trajectory; notably, it incorporates sparse attention to prevent irrelevant brain region interactions from overwhelming useful information. Finally, \(\hat{x}_{HF}\) is the generated high-field MRI volume.

\subsection{Unified Modeling via Video Super-Resolution Prior}
Acquiring paired high-field MRI data is expensive and challenging. To overcome this data scarcity, we propose a dual approach: unified modeling to maximize data utilization, and an external prior to bypass training from scratch.

\textbf{Unified Modeling.} Lower-field MRI degradation patterns are highly consistent across modalities (e.g., T1, T2, FLAIR) and field transitions (e.g., 64mT-to-3T, 3T-to-7T). Consolidating these tasks into a single framework conditioned on modality and transition acts as implicit data augmentation. This enables the joint learning of shared enhancement features—such as noise suppression and detail recovery—drastically reducing the need for massive paired data per task.

\textbf{Video Super-Resolution Prior.} Training 3D generative models from scratch is heavily data-dependent. Because 3D MRI and 2D video share analogous spatiotemporal structures and enhancement goals, we introduce a pre-trained video prior to provide rich, generalized structural representations. Specifically, we initialize our framework with FlashVSR \cite{flashvsr}. To efficiently adapt this powerful generative prior to 3D MRI, we freeze the encoder and decoder, and employ Low-Rank Adaptation (LoRA) to fine-tune the main body.

\subsection{Field-Aware Spectral Rectification Mechanism (FASRM)}

To overcome the spectral bias and over-smoothing in traditional flow-matching, we propose a field-aware spatial-frequency loss (FASFL) within a field-aware spectral rectification mechanism (FASRM). Unlike uniform or task-agnostic spectral penalties, FASFL dynamically modulates frequency-domain optimization based on the physical properties of the source and target magnetic fields.

Specifically, FASRM introduces a field-conditioned spectral rectification scheme in the frequency domain to address the fundamentally different physical constraints of diverse enhancement tasks. For example, in extreme low-to-high field transitions (e.g., 64mT-to-3T), high-frequency features of the target high-field images lack corresponding structural hints in the low-field source. Over-penalizing the high-frequency band would force the model to blindly hallucinate these structures. Thus, FASRM automatically relaxes high-frequency constraints to prioritize structural fidelity. As another example, in high-to-ultra-high field transitions (e.g., 3T-to-7T), target images frequently suffer from inherent physical artifacts, such as B1 inhomogeneity, which predominantly manifest in the low-frequency spectrum. To prevent the model from memorizing and reproducing these artifacts, FASRM suppresses the low-frequency learning weights. By explicitly embedding these field-specific physical priors, our method effectively mitigates both hallucinations and artifact propagation.

In summary, the unified objective, which simultaneously enforces spatial fidelity and spectral rectification, is defined as:
\begin{align}
\mathcal{L}_{FASFL} &= \| v_\theta - v \|_1 \nonumber\\
&+ \lambda_{freq} \sum_{b \in \mathcal{B}} \frac{w_b}{|M_b|} \left\| M_b \odot \left( |F(v_\theta) - F(v)|^\alpha \odot |F(v_\theta) - F(v)|^2 \right) \right\|_1, 
\end{align}
where \(v_\theta\) and \(v\) are the predicted and target velocity fields, and \(F(\cdot)\) denotes the Fast Fourier Transform (FFT). \(\mathcal{B} = \{low, mid, high\}\) denotes the partitioned frequency bands with binary masks \(M_b\), where $|M_b|$ represents the number of elements in the mask. Crucially, the field-conditioned weights \(w_b\) are designed for adapting to the specific field transition, \(\alpha\) modulates the degree of focus on challenging spectra, and \(\lambda_{freq}\) controls the strength of spectral rectification.

\begin{comment}

In summary, the overall training objective of our framework comprises a spatial reconstruction loss and the proposed physics-guided frequency loss. In the spatial domain, we apply an \(L_1\) loss to constrain the predicted velocity field \(v_\theta\) against the target velocity field \(v\):
\[
\mathcal{L}_{spat} = \| v_\theta - v \|_1
\]
In the frequency domain, we compute the Focal Frequency Loss (FFL) across the partitioned bands. Let \(F_\theta\) and \(F\) denote the 3D Fast Fourier Transform (FFT) representations of the predicted and target velocities, respectively. The frequency loss is defined as:
\[
\mathcal{L}_{freq} = \sum_{b \in \{low, mid, high\}} w_b \frac{1}{|M_b|} \sum_{\mathbf{k} \in M_b} \left( |F_\theta(\mathbf{k}) - F(\mathbf{k})|^\alpha \cdot \| F_\theta(\mathbf{k}) - F(\mathbf{k}) \|^2 \right)
\]
where \(M_b\) represents the binary mask for frequency band \(b\), \(\mathbf{k}\) denotes the 3D frequency coordinates, \(w_b\) is the task-specific base coefficient discussed above, and \(\alpha\) is the focal scaling factor that dynamically scales the weight based on the generation difficulty. Finally, the total loss function is formulated as a weighted sum of both domains:
\[
\mathcal{L}_{total} = \lambda_{spat} \mathcal{L}_{spat} + \lambda_{freq} \mathcal{L}_{freq}
\]
where \(\lambda_{spat}\) and \(\lambda_{freq}\) are hyperparameters balancing the spatial and spectral constraints.

\end{comment}

\section{Experiments}
\subsection{Dataset and Experimental Details}
%We evaluated two cross-field MRI translation tasks using five datasets (Table \ref{Dataset}): 64mT-to-3T (ULF-EnC, Leiden, KCL) and 3T-to-7T (UNC, BNU), each split 8:2 for training and testing. Preprocessing included MONAI-based intensity normalization (clipping to the 0.5th--99.5th percentiles), spatial registration (newly registered datasets will be open-sourced), resampling to a 1mm z-resolution, and cropping to 256 \(\times\) 256 \(\times\) 160.

%We evaluate two cross-field MRI translation tasks using five datasets (Table \ref{Dataset}): 64mT-to-3T (ULF-EnC, Leiden, KCL) and 3T-to-7T (UNC, BNU), each split 8:2 for training and testing. Crucially, we perform rigorous spatial registration on previously unaligned data, organizing and publicly releasing a comprehensive benchmark that is an order of magnitude larger than existing datasets. Further preprocessing includes MONAI-based intensity normalization (clipping to the 0.5th--99.5th percentiles), resampling to a 1mm z-resolution, and cropping to \(256 \times 256 \times 160\).

%Empirically, we set \(\lambda_{spat} = 1\), \(\lambda_{freq} = 0.1\), and \(\alpha = 1.0\). Following our physical priors, the frequency band weights \(w_b\) (low, mid, high) were set to \([0.2, 0.5, 0.3]\) for 64mT-to-3T, and \([0.1, 0.3, 0.6]\) for 3T-to-7T. The model was implemented in PyTorch and trained on an NVIDIA RTX A6000 GPU for 400K iterations with a batch size of 1. We utilized the Adam optimizer (\(\beta_1 = 0.5\), \(\beta_2 = 0.999\)) with an initial learning rate of \(10^{-4}\), which linearly decays after the first 200K iterations.

To overcome the limitations of small-scale, single-center studies, we curate a comprehensive multi-center, multi-task MRI dataset sourced from five institutions. Featuring rigorous preprocessing and precise cross-field registration, this dataset enables the systematic evaluation of both $64\text{mT} \rightarrow 3\text{T}$ and $3\text{T} \rightarrow 7\text{T}$ enhancements. As detailed in Table \ref{Dataset}, all experiments utilize a uniform 8:2 train-test split across each center. Preprocessing includes MONAI intensity normalization (0.5th--99.5th percentiles), 1mm z-resampling, and \(256 \times 256 \times 160\) resizing.

We set \(\lambda_{freq}=0.1\), \(\alpha=1.0\), and frequency weights \(w_b\) to \([0.2, 0.5, 0.3]\) (64mT-to-3T) and \([0.1, 0.3, 0.6]\) (3T-to-7T). Implemented in PyTorch, the model was trained on an RTX A6000 (batch size 1, 1,000 iterations) using Adam (\(\beta_1=0.5\), \(\beta_2=0.999\), initial \(lr=10^{-4}\) with linear decay after 500).

\begin{comment}
This study utilizes a total of five datasets for two cross-field MRI translation tasks: the 64mT-to-3T task includes three datasets (ULF-EnC, Leiden, and KCL), while the 3T-to-7T task includes two datasets (UNC and BNU), as detailed in Table 1. Each dataset was randomly divided into training and testing sets at an 8:2 ratio. To eliminate intensity variations, we normalized all images using MONAI by clipping and scaling to the 0.5th and 99.5th percentiles. Furthermore, we performed spatial registration on the originally unregistered datasets, and these registered datasets will be made publicly available. Additionally, all MRI volumes were resampled to a 1mm resolution in the z-direction and cropped to a uniform spatial dimension of 256 \(\times\) 256 \(\times\) 160. 

In addition, we empirically set \(\lambda_{spat} = 1\), \(\lambda_{freq} = 0.1\), and the focal parameter \(\alpha = 1.0\). Following our physical priors, the field-conditioned weights \(w_b\) for the low, mid, and high-frequency bands were set to \([0.2, 0.5, 0.3]\) for the 64mT-to-3T task, and \([0.1, 0.3, 0.6]\) for the 3T-to-7T task. Moreover, our model is implemented in Python (PyTorch) and runs on an NVIDIA RTX A6000 GPU. The training utilizes the Adam optimizer (\(\beta_1 = 0.5\), \(\beta_2 = 0.999\)), with an initial learning rate of 0.0001, which linearly decays after half of the total iterations (400K iterations). The batch size is set to 1.
\end{comment}

\begin{table*}[!tp]
\centering
\caption{\textbf{Dataset overview.} For unregistered MRI pairs, we perform registration.}
\label{Dataset}
\resizebox{0.6\textwidth}{!}{
\begin{NiceTabular}{c|c|c|c|ccc}
\shline
\textbf{Task} & \textbf{Dataset} & \textbf{Modalities} & \textbf{Registered} & \textbf{Total} & \textbf{Train} & \textbf{Test} \\ 
\midrule
\multirow{3}{*}{\textbf{64mT-to-3T}} & ULF-EnC \cite{islam2023improving} & T1, T2, FLAIR & Yes & 50 & 40 & 10 \\ 
                                     & Leiden \cite{vandenbroek2025paired} & T1, T2, FLAIR & No  & 10 & 8  & 2  \\ 
                                     & KCL \cite{vasa2025ultra}            & T1, T2        & No  & 23 & 18 & 5  \\ 
\hline
\multirow{2}{*}{\textbf{3T-to-7T}}   & UNC \cite{chen2023paired}           & T1, T2        & Yes & 10 & 8  & 2  \\ 
                                     & BNU \cite{chu2025paired}            & T1, T2        & No  & 20 & 16 & 4  \\ 
\shline
\end{NiceTabular} 
}
\end{table*}

\subsection{Comparative Experiments}

\begin{figure}[t]

	\centerline{\includegraphics[width=\columnwidth]{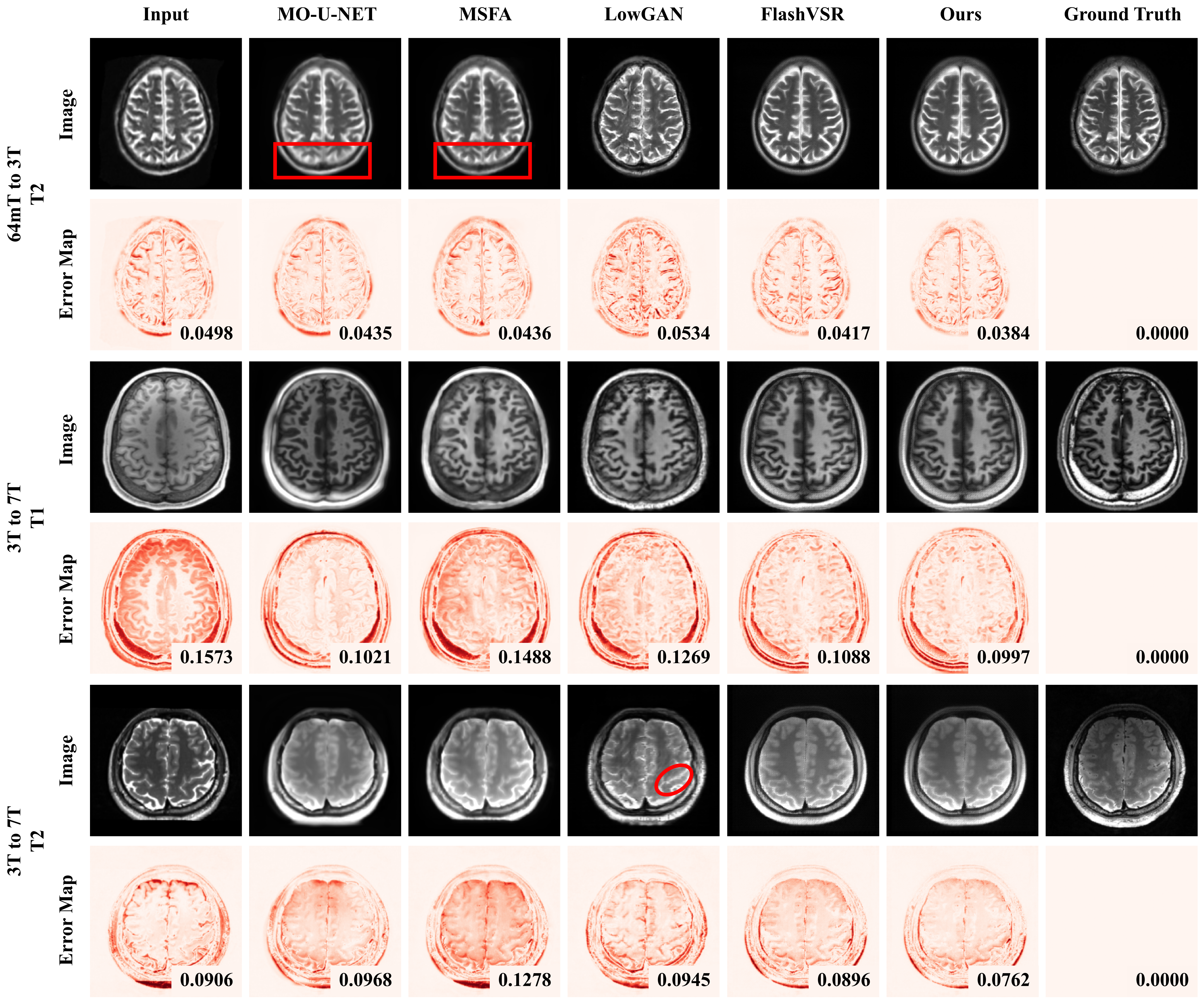}}
    \caption{\textbf{Visual comparison with competing methods.} Error maps show the absolute difference from the ground truth, where darker colors indicate larger errors (mean pixel error at bottom right). Red annotations  highlight defects in competing methods.}
	\label{Comparative_Experiments}
\end{figure}

In our comparative experiments, we evaluate our method against several baselines, including the U-Net-based MO-U-NET \cite{MO-U-NET}, GAN-based MSFA \cite{MSFA} and LowGAN \cite{lowgan}, and diffusion-based FlashVSR \cite{flashvsr}. As observed in Fig. \ref{Comparative_Experiments}, MO-U-NET and MSFA generate highly blurred images, particularly in the red boxes of the 64mT-to-3T task. Meanwhile, LowGAN mistakenly generates normal brain tissue as a bright white hyperintense artifact, as shown in the red circle of the T2 3T-to-7T task. In addition, although FlashVSR produces better results than the previous methods, its error map still exhibits darker colors, especially in high-frequency details. In contrast, our method successfully addresses most of these issues, recovering sharp tissue boundaries and rich textures that are highly consistent with the ground truth, which is further validated by having the lowest residual values in the error maps. Quantitatively, as shown in Tables \ref{comp-1} and \ref{comp-2}, our method achieves the best performance across the vast majority of evaluation metrics. This comprehensive superiority in both visual fidelity and specific numerical gains fully demonstrates the effectiveness of our proposed approach.

\begin{comment}
%  chwang updated
\begin{table*}[!tp]
\centering
\caption{\textbf{Quantitative comparison our \methodname and other competing methods  on the 64mT-to-3T task for T1, T2, and FLAIR modalities.} 
NRMSE ($\downarrow$) and SSIM ($\uparrow$) are in \%, PSNR ($\uparrow$) in dB, and LPIPS ($\downarrow$) is unitless ($\uparrow$/$\downarrow$: higher/lower is better).
Best results are \textbf{bolded}.}
\label{comp-1}
{\tiny
\begin{NiceTabular*}{1.\textwidth}{@{\extracolsep{\fill}} l|cccc|cccc|cccc @{}}

\shline
\multirow{2}{*}{Methods} & \multicolumn{4}{c}{\textit{T1}} & \multicolumn{4}{c}{\textit{T2}} & \multicolumn{4}{c}{\textit{FLAIR}} \\
& NRMSE  & PSNR & SSIM  & LPIPS  & NRMSE  & PSNR  & SSIM  & LPIPS & NRMSE  & PSNR  & SSIM  & LPIPS  \\
\midrule
MO-U-NET(2025) & 11.4 & 18.98 & 52.9 & 0.28 & \; 9.5  & 20.58 & 59.2 & 0.23 & 10.9 & 19.32 & 53.3 & 0.25 \\
MSFA(2024)     & 11.3 & 19.05 & 67.9 & 0.26 & \; 9.9  & 20.27 & 71.6 & 0.22 & 10.9 & 19.31 & 65.6 & 0.24 \\
LowGAN(2025)   & 12.8 & 18.03 & 66.6 & 0.19 & 11.4 & 18.98 & 65.8 & 0.20 & 14.1 & 17.02 & 63.2 & \textbf{0.20} \\
FlashVSR(2026) & 12.1 & 18.41 & 63.4 & 0.30 & 12.0 & 18.54 & 66.3 & 0.23 & 15.1 & 16.61 & 60.6 & 0.30 \\
\hline
\rowcolor{cyan!10}\textbf{\methodname(Ours)} & \textbf{10.5} & \textbf{19.75} & \textbf{72.8} & \textbf{0.19} & \;\textbf{8.4} & \textbf{21.57} & \textbf{74.9} & \textbf{0.16} & \;\textbf{9.5} & \textbf{20.50} & \textbf{71.5} & \textbf{0.20} \\
\shline
\end{NiceTabular*} 
}
\end{table*}
\end{comment}

\begin{table*}[!tp]
\centering
\caption{\textbf{Quantitative comparison of our \methodname and other competing methods on the 64mT-to-3T task for T1, T2, and FLAIR modalities.} 
NRMSE ($\downarrow$) and SSIM ($\uparrow$) are in \%, PSNR ($\uparrow$) in dB, and LPIPS ($\downarrow$) is unitless ($\uparrow$/$\downarrow$: higher/lower is better).
Best and second-best results are \textbf{bolded} and \underline{underlined}, respectively.}
\label{comp-1}
{\tiny
\begin{NiceTabular*}{1.\textwidth}{@{\extracolsep{\fill}} l|cccc|cccc|cccc @{}}

\shline
\multirow{2}{*}{Methods} & \multicolumn{4}{c}{T1} & \multicolumn{4}{c}{T2} & \multicolumn{4}{c}{FLAIR} \\
& NRMSE  & PSNR & SSIM  & LPIPS  & NRMSE  & PSNR  & SSIM  & LPIPS & NRMSE  & PSNR  & SSIM  & LPIPS  \\
\midrule
MO-U-NET(2025) & 11.4 & 18.98 & 52.9 & 0.28 & \underline{\; 9.5}  & \underline{20.58} & 59.2 & 0.23 & \underline{10.9} & \underline{19.32} & 53.3 & 0.25 \\
MSFA(2024)     & \underline{11.3} & \underline{19.05} & \underline{67.9} & \underline{0.26} & \; 9.9  & 20.27 & \underline{71.6} & 0.22 & \underline{10.9} & 19.31 & \underline{65.6} & \underline{0.24} \\
LowGAN(2025)   & 12.8 & 18.03 & 66.6 & \textbf{0.19} & 11.4 & 18.98 & 65.8 & \underline{0.20} & 14.1 & 17.02 & 63.2 & \textbf{0.20} \\
FlashVSR(2026) & 12.1 & 18.41 & 63.4 & 0.30 & 12.0 & 18.54 & 66.3 & 0.23 & 15.1 & 16.61 & 60.6 & 0.30 \\
\hline
\rowcolor{cyan!10}\textbf{\methodname(Ours)} & \textbf{10.5} & \textbf{19.75} & \textbf{72.8} & \textbf{0.19} & \;\textbf{8.4} & \textbf{21.57} & \textbf{74.9} & \textbf{0.16} & \;\textbf{9.5} & \textbf{20.50} & \textbf{71.5} & \textbf{0.20} \\
\shline
\end{NiceTabular*} 
}
\end{table*}

\begin{table*}[!tp]
\centering
\caption{\textbf{Quantitative comparison of our \methodname and other competing methods on the 3T-to-7T task for T1 and T2 modalities.} 
NRMSE ($\downarrow$) and SSIM ($\uparrow$) are in \%, PSNR ($\uparrow$) in dB, and LPIPS ($\downarrow$) is unitless ($\uparrow$/$\downarrow$: higher/lower is better).
Best and second-best results are \textbf{bolded} and \underline{underlined}, respectively.
}
\label{comp-2}
{\tiny
\begin{NiceTabular*}{.68\textwidth}{ l|cccc|cccc}
\shline
\multirow{2}{*}{Methods} & \multicolumn{4}{c}{T1} & \multicolumn{4}{c}{T2} \\
& NRMSE & PSNR & SSIM  & LPIPS  & NRMSE  & PSNR & SSIM  & LPIPS \\
\midrule
MO-U-NET (2025) & 11.9 & 18.51 & 60.8 & 0.28 & \,\;8.6  & 21.32 & 59.7 & 0.27 \\
MSFA (2024)    & \underline{10.5} & \underline{19.66} & \textbf{66.8} & \underline{0.23} & \,\;\underline{8.1}  & \underline{21.82} & \underline{65.0} & \underline{0.24} \\
LowGAN (2025)  & 15.4 & 16.26 & 54.0 & 0.25 & 10.1 & 19.97 & 56.0 & 0.25 \\
FlashVSR  (2026) & 11.8 & 18.63 & 63.0 & \underline{0.23} & \,\;9.5  & 20.49 & 58.6 & 0.24 \\
\hline
\rowcolor{cyan!10}\textbf{\methodname(Ours)} & \,\;\textbf{9.9}  & \textbf{20.12} & \underline{66.1} & \textbf{0.21} & \,\;\textbf{7.5}  & \textbf{22.55} & \textbf{72.1} & \textbf{0.18} \\
\shline
\end{NiceTabular*} 
}
\end{table*}

\subsection{Ablation Experiments}

\begin{table*}[!tp]
\centering
\caption{\textbf{Quantitative ablation on multi-modality unification.} % We compare training separate models for individual modalities (T1, T2, and FLAIR) versus a single unified model for all modalities. 
NRMSE ($\downarrow$) and SSIM ($\uparrow$) are in \%, PSNR ($\uparrow$) in dB, and LPIPS ($\downarrow$) is unitless ($\uparrow$/$\downarrow$: higher/lower is better). Best results are \textbf{bolded}.
}
\label{abl-1}
{\tiny
\begin{NiceTabular*}{1.\textwidth}{@{\extracolsep{\fill}} l|cccc|cccc|cccc @{}}
\shline
\multirow{2}{*}{Methods} & \multicolumn{4}{c}{T1} & \multicolumn{4}{c}{T2} & \multicolumn{4}{c}{FLAIR} \\
& NRMSE  & PSNR & SSIM  & LPIPS  & NRMSE  & PSNR  & SSIM  & LPIPS & NRMSE  & PSNR  & SSIM  & LPIPS  \\
\midrule
\multicolumn{13}{c}{\textit{Task: 64mT-to-3T}} \\
\midrule
T1 Only    & 12.1 & 18.41 & 63.4 & 0.30 & - & - & - & - & - & - & - & - \\
T2 Only    & - & - & - & - & 12.0 & 18.54 & 66.3 & 0.23 & - & - & - & - \\
FLAIR Only & - & - & - & - & - & - & - & - & 15.1 & 16.61 & 60.6 & 0.30 \\
\hline
Unified modalities & 11.7 & 18.71 & 65.1 & 0.28 & \; 9.4 & 20.58 & 68.6 & 0.23 & 11.1 & 19.17 & 64.5 & 0.30 \\
\quad + Unified tasks & 10.7 & 19.06 & 72.1 & 0.20 & \; 8.9 & 21.05 & 72.8 & 0.18 & 11.9 & 19.06 & 66.7 & 0.24 \\
\rowcolor{cyan!10}\textbf{\quad\quad + FASRM (Ours)} & \textbf{10.5} & \textbf{19.75} & \textbf{72.8} & \textbf{0.19} & \;\textbf{8.4} & \textbf{21.57} & \textbf{74.9} & \textbf{0.16} & \;\textbf{9.5} & \textbf{20.50} & \textbf{71.5} & \textbf{0.20} \\
\midrule
\multicolumn{13}{c}{\textit{Task: 3T-to-7T}} \\
\midrule
T1 Only & 11.8 & 18.63 & 63.0 & 0.23 & - & - & - & - & - & - & - & - \\
T2 Only & - & - & - & - & \; 9.5 & 20.49 & 58.6 & 0.24 & - & - & - & - \\
\hline
Unified modalities & 11.0 & 19.25 & 64.4 & \textbf{0.21} & \; 9.0 & 20.97 & 61.1 & 0.21 & - & - & - & - \\
\quad + Unified tasks  & 10.6 & 19.53& 65.8 & \textbf{0.21} & \; 9.2 & 20.88 & 66.6 & 0.19 & - & - & - & - \\
\rowcolor{cyan!10}\textbf{\quad\quad + FASRM (Ours)} & \,\;\textbf{9.9}  & \textbf{20.12} & \textbf{66.1 }& \textbf{0.21} & \,\;\textbf{7.5}  & \textbf{22.55} & \textbf{72.1} & \textbf{0.18} & - & - & - & - \\
\shline
\end{NiceTabular*} 
}
\end{table*}

%Building upon modality unification, we further compare task-specific networks (e.g., separate models for 64mT-to-3T and 3T-to-7T) against a single unified cross-task model. As demonstrated in the Table \ref{abl-1}, the unified model significantly outperforms its task-specific counterparts. This substantial improvement can be attributed to the shared underlying mechanisms of field-strength enhancement. Regardless of whether the transition is from 64mT-to-3T or from 3T-to-7T, the core enhancement process involves highly analogous mappings: suppressing noise and recovering high-frequency anatomical details. By jointly training on both tasks, the network learns a more robust, generalized prior for magnetic field upscaling, allowing it to effectively transfer structural representations across different field-strength domains.

%Finally, we validate the effectiveness of the proposed Field-Aware Spatial-Frequency Loss (FASFL). As illustrated in the Fig. \ref{abl-fig}, the absence of FASFL leads to poor reconstruction of high-frequency details, resulting in noticeably blurred generated images, as seen in the 64mT-to-3T results. Furthermore, as highlighted by the red box in the 3T-to-7T task, the model without FASFL generates anatomical structures that are inconsistent with the input image. By incorporating FASFL, these issues are effectively resolved, yielding sharper and anatomically faithful results. The quantitative metrics in the Table \ref{abl-1} further demonstrate the performance improvements brought by FASFL, fully confirming the effectiveness of our proposed approach.

We first compare a unified multi-modality model against modality-specific models. As shown in Table \ref{abl-1}, the unified model consistently outperforms individual models across 64mT-to-3T and 3T-to-7T. Since lower-field degradation patterns are similar across modalities, joint training acts as data augmentation. Additionally, a unified framework reduces storage and streamlines clinical workflows.

Building upon this, a single unified cross-task model significantly surpasses task-specific networks (Table \ref{abl-1}). Since field enhancement inherently involves analogous mappings---such as noise suppression and high-frequency recovery---joint training enables the network to learn a robust, generalized prior, effectively transferring structural representations across different field-strength domains.

\begin{figure}[t]
	\centerline{\includegraphics[width=\columnwidth]{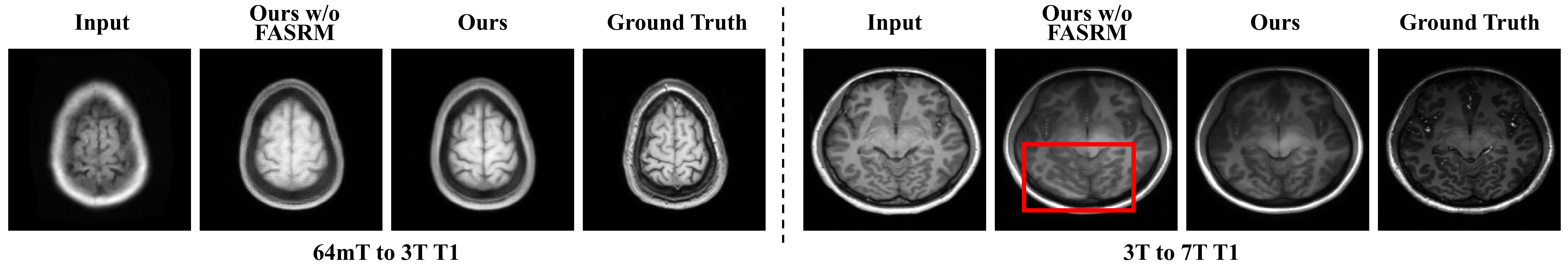}}
    \caption{\textbf{Visualization of ablation experiments on FASRM.} Results for both 64mT-to-3T and 3T-to-7T tasks demonstrate the module's effectiveness. Red boxes highlight severe defects in the baselines without FASRM.}
    \label{abl-fig}
\end{figure}

Finally, we validate the proposed FASRM. As illustrated in Fig. \ref{abl-fig}, omitting FASRM leads to poor high-frequency reconstruction, yielding blurred images in the 64mT-to-3T task and anatomically inconsistent structures in the 3T-to-7T task. Incorporating FASRM effectively resolves these issues, producing sharper, anatomically faithful results. The quantitative metrics in Table \ref{abl-1} further corroborate the performance improvements brought by FASRM.

\section{Conclusion}
%In this paper, we introduce UniField, a unified framework that significantly advances MRI field-strength enhancement by overcoming data scarcity and generalization bottlenecks. We construct the largest registered multi-field dataset to date and leverage video super-resolution priors alongside shared degradation patterns to maximize data efficiency. Furthermore, our novel field-aware spectral rectification mechanism effectively resolves the spectral bias inherent in generative models, successfully restoring sharp high-frequency anatomical details. Building upon this robust foundation, our future work will focus on expanding this paradigm to integrate a wider range of anatomical organs, diverse magnetic field strengths, and various disease types, ultimately paving the way for universally accessible, high-quality clinical diagnostics.
In this paper, we introduce UniField, a unified framework that advances MRI field-strength enhancement by overcoming data scarcity and generalization bottlenecks. We construct the largest registered multi-field dataset to date and leverage video super-resolution priors with shared degradation patterns to maximize data efficiency. Furthermore, our novel field-aware spectral rectification mechanism resolves the spectral bias inherent in generative models, restoring sharp high-frequency anatomical details. Future work will expand this paradigm to diverse anatomical organs, magnetic field strengths, and disease types, paving the way for universally accessible, high-quality clinical diagnostics.

\vspace{1em} % 增加垂直间距，1em代表当前字体大小的一个字宽的高度

\noindent \textbf{Acknowledgement.} This work was supported by NSFC/RGC Joint Research Scheme N\_CUHK4126/25, Innovation and Technology Fund Mainland-Hong Kong Joint Funding Scheme MHP/173/24.

\vspace{1em} % 增加垂直间距，1em代表当前字体大小的一个字宽的高度

\noindent \textbf{Disclosure of Interests.} The authors have no competing interests to declare that are
relevant to the content of this article.

\bibliographystyle{splncs04}
\bibliography{ref}
\end{sloppypar}
\end{document}